\documentclass[11pt]{article}
\usepackage[preprint]{acl}
\usepackage{times}
\usepackage{latexsym}
\usepackage[T1]{fontenc}
\usepackage[utf8]{inputenc}
\usepackage{microtype}
\usepackage{inconsolata}
\usepackage{graphicx}
\usepackage{amsmath}
\usepackage{amssymb}
\usepackage{booktabs}
\usepackage{multirow}
\usepackage{placeins}

\title{Attention-Guided Layer Selection for Contrastive Decoding in Large Language Models}

\author{
  Yusuke Sakai, 
  Natthawut Kertkeidkachorn, 
  Kiyoaki Shirai \\
  Japan Advanced Institute of Science and Technology \\
  \texttt{\{yusuke.sakai,natt,kshirai\}@jaist.ac.jp}
}

\begin{document}
\maketitle

\begin{abstract}
% DoLa は単一モデル内の最終層と浅い層の出力分布を対比することで事実性を改善する contrastive decoding 手法であり、出力分布間の Jensen--Shannon Divergence に基づいて対比層を動的に選択する。本研究では、出力分布ではなく自己注意を層選択シグナルとして用いる3つの戦略（Attention-JSD, Attention-Entropy-Max, Attention-Entropy-Min）を提案する。

Contrastive decoding methods such as DoLa improve the factuality of Large Language Models (LLMs) by contrasting the output distributions of mature and premature layers. However, DoLa’s dynamic layer selection relies solely on divergences in output vocabulary distributions. In this work, we propose three attention-guided strategies—Attention-JSD, Attention-Entropy-Max, and Attention-Entropy-Min—that leverage structural information carried by internal self-attention mechanisms as a signal for layer selection. Experimental results on TruthfulQA demonstrate that our strategies, particularly Attention-JSD and Attention-Entropy-Min, consistently outperform the original DoLa. We observe significant gains on multi-answer metrics (MC2 and MC3), suggesting that attention distributions can provide a more sensitive signal for resolving factual knowledge than output vocabulary distributions. 

\end{abstract}

\section{Introduction}
\label{sec:introduction}

% 大規模言語モデル(LLM)は、多様な自然言語処理タスクにおいて高い性能を示している一方で、事実と異なる内容をもっともらしく生成するハルシネーションの問題を抱えている。特に、オープンドメイン質問応答や長文生成のようなタスクでは、モデルが訓練データに基づかない根拠のない主張を生成することが、実用上の大きな障害となっている。

Large language models (LLMs) have achieved strong performance across a wide range of natural language processing tasks, yet they remain prone to hallucination, i.e., generating plausible but factually incorrect content \citep{Hal-survey}. This issue is particularly problematic in tasks such as open-domain question answering and long-form generation, where producing unsupported claims that are not grounded in the training data poses a major obstacle to real-world deployment.

% この問題に対する推論時の対処法の一つとして、contrastive decoding \citep{li2022contrastive} がある。contrastive decoding は、単一モデルの出力分布のみに依存せず、複数の分布を対比させることで、生成における望ましくない傾向を相対的に抑制し、より好ましい候補を選好する枠組みである。たとえば、Liら.\ \citep{li2022contrastive} は、大規模モデル（expert）と小規模モデル（amateur）の対数確率の差を目的関数とし、さらに大規模モデルの確信度に基づく妥当性制約を導入することで、反復や話題逸脱といったオープンエンド生成の劣化を抑えつつ、流暢さと一貫性を維持した生成を目指す。

One inference-time approach to mitigating this issue is contrastive decoding \citep{li2022contrastive}. Contrastive decoding is a framework that does not rely solely on the output distribution of a single model; instead, it contrasts multiple distributions to relatively suppress undesirable generation tendencies and to prefer more desirable candidates. For instance, Li et al.\ \citep{li2022contrastive} define an objective based on the difference between the log-probabilities of a large model (expert) and a small model (amateur), and introduce a plausibility constraint based on the expert model’s confidence, aiming to reduce degradations in open-ended generation such as repetition and topic drift while maintaining fluency and coherence.

% DoLa (Decoding by Contrasting Layers) \citep{chuang2024dola} は、この対比の発想を単一モデル内部へと拡張し、最終層（mature layer）と初期層（premature layer）のlogits を対比することで、追加学習なしに事実性の改善を図る手法である。DoLa は、候補となる premature 層の集合に対して、最終層との Jensen--Shannon Divergence (JSD) が最大となる層をトークンごとに動的に選択し、TruthfulQA \citep{lin2022truthfulqa} や FACTOR \citep{muhlgay2024factor} などのベンチマークで有効性を示している。さらに、別モデルや外部シグナルを必要とせず、単一モデルの内部情報のみで完結する点は、推論時に導入しやすい実用上の利点である。一方で、DoLa における層選択は出力分布（logits）間の差異に基づいており、モデル内部の注意機構が持つ情報を層選択のシグナルとしては直接には利用していない。

DoLa (Decoding by Contrasting Layers) \citep{chuang2024dola} extends this contrastive idea within a single model by contrasting the logits of the final (mature) layer and an early (premature) layer, thereby improving factuality without additional training. Given a set of candidate premature layers, DoLa dynamically selects, at each token, the layer that maximizes the Jensen--Shannon divergence (JSD) from the final layer, and demonstrates effectiveness on benchmarks such as TruthfulQA \citep{lin2022truthfulqa} and FACTOR \citep{muhlgay2024factor}. Moreover, DoLa is self-contained within a single model, requiring neither an auxiliary model nor external signals, which makes it a practical plug-in method for inference-time adoption. However, DoLa’s layer selection is driven by differences in output distributions (logits) and does not directly leverage information carried by the model’s internal attention mechanisms as a selection signal.

% Transformer の内部には、出力 logits 以外にも、自己注意機構（self-attention）や中間表現など、多様で豊富な情報源が存在する。近年は、隠れ状態のクロス層エントロピー変化を用いた事実性向上 \citep{wu2025end}、検索拡張生成において文脈関連度に応じてコントラスト強度を適応的に調整する手法 \citep{kim2024acd,shi2024trusting}、注意パターンそのものを幻覚検出に利用する研究 \citep{chuang2024lookback} など、出力分布以外のシグナルを活用する試みが報告されている。これらの知見は、LLM の内部に生成制御に有用な相補的手がかりが複数存在しうることを示唆する。一方で、コントラスティブなデコーディング手法の多くは、logits や隠れ状態から得られるスカラー統計量に主に依存しており、注意分布が持つ構造的情報を、トークン単位の層選択や対比設計の動的シグナルとして系統的に取り込む試みは限定的である。したがって、DoLa の層選択に注意情報を統合することは、単一モデル内で完結するという利点を保ちつつ、より適切な対比層の選択を通じた事実性向上につながる可能性がある。

Transformers provide diverse internal signals beyond output distributions, including self-attention mechanisms and intermediate representations. Recent work has explored leveraging such signals, for example, using cross-layer entropy changes of hidden states \citep{wu2025end}, adaptively tuning contrast strength in retrieval-augmented generation based on context \citep{kim2024acd,shi2024trusting}, and detecting hallucinations from attention patterns \citep{chuang2024lookback}. 
Nevertheless, contrastive decoding methods still often rely on scalar statistics derived from logits or hidden states, and systematic attempts to use the structural information in attention distributions as token-level signals for layer selection remain limited.

% 本研究では、注意機構が事実知識が現れる層を特定するためのシグナルになりうるという仮説を立てる。具体的には、DoLaの枠組みを維持しつつ、内部注意分布に基づいて premature層を選択する3つの戦略を導入する。

In this study, we hypothesize that attention mechanisms can serve as a signal for identifying layers where factual knowledge emerges. Specifically, while keeping the DoLa framework intact, we introduce three strategies for selecting premature layers based on internal attention distributions.

Our contributions are as follows:
\begin{itemize}
    \item We propose three attention-guided strategies---Attention-JSD, Attention-Entropy-Max, and Attention-Entropy-Min---that leverage structural information derived from self-attention mechanisms as dynamic signals for layer selection, moving beyond the conventional reliance on output vocabulary distributions.
    \item TruthfulQA evaluations show that our proposed methods (especially Attention-JSD and Attention-Entropy-Min) outperform existing approaches, achieving substantial gains particularly on the multiple-correct-answer metrics (MC2/MC3). In addition, on FACTOR, our methods improve over the baseline in many settings and achieve performance comparable to, or in some cases better than, DoLa.
    \item Through visualization of layer selection patterns and head-wise analysis, we identify that specific attention heads carry distinct signals conducive to factuality discrimination, thereby revealing potential avenues for further optimization in contrastive decoding design.
\end{itemize}

\section{Related Work}
\label{sec:related}

\paragraph{Hallucinations in Large Language Models.}
% LLMにおけるハルシネーションは、検証可能な事実と矛盾する内容を生成する事実性ハルシネーション(factuality hallucination)と、与えられたコンテキストや指示から逸脱する忠実性ハルシネーション(faithfulness hallucination)に大別される。いずれの形態も、対話システム、検索拡張生成、エージェントなど幅広い応用において深刻な問題となっている。~\citep{ji2023survey,Hal-survey}。この問題を軽減するために、人間のフィードバックからの強化学習(RLHF)~\citep{ouyang2022training}、推論時の自己一貫性チェック ~\citep{wang2023selfconsistency,manakul2023selfcheckgpt}、マルチエージェント討論~\citep{du2024debate}など、様々なアプローチが提案されている。これらの手法は通常、追加の訓練、外部知識~\citep{lewis2020rag}、または複数回の推論を必要とする。本研究は、既存モデルのパラメータを変更せず、単一の前向き計算で実現可能な手法に焦点を当てる。

LLM hallucinations can be broadly categorized into factuality hallucinations, which generate content that contradicts verifiable facts, and faithfulness hallucinations, which deviate from the given context or instructions. Both forms pose serious challenges across a wide range of applications, including dialogue systems, retrieval-augmented generation (RAG), and agents~\citep{ji2023survey,Hal-survey}. To mitigate this issue, various approaches have been proposed, such as reinforcement learning from human feedback (RLHF)~\citep{ouyang2022training}, self-consistency checks at inference time~\citep{wang2023selfconsistency,manakul2023selfcheckgpt}, and multi-agent debate~\citep{du2024debate}. These methods typically require additional training, access to external knowledge~\citep{lewis2020rag}, or multiple inference passes. In contrast, this study focuses on methods that do not modify the parameters of existing models and can be realized with a single forward computation.

\paragraph{Layer-wise Knowledge in Transformers.}
% Transformer ベースモデルが情報を層ごとに階層的に符号化することは、多くの分析研究により示されている~\citep{tenney2019bert,jawahar2019bert}。例えば BERT では、下層で表層的特徴や局所的な統語情報が捉えられ、中層から上層にかけて意味的・文脈的情報がより顕在化する傾向が報告されている~\citep{jawahar2019bert}。また、言語理解の各段階に対応する情報が層方向に順に現れるという観察もある~\citep{tenney2019bert}。 こうした観察に基づき、モデル内部のどの計算が事実連想を媒介するかを同定し、その局所的な計算を編集することで知識を更新する手法が提案されている~\citep{meng2022rome,meng2023memit}。さらに、層や注意機構に現れる内部表現を推論時に直接介入して望ましい性質を引き出す試みとして、特定の注意ヘッドの活性を操作して truthfulness を改善する Inference-Time Intervention (ITI) も提案されている~\citep{li2023inferencetime}。

Transformer-based models have been shown in many analysis studies to encode information hierarchically across layers~\citep{tenney2019bert,jawahar2019bert}. For example, in BERT, lower layers tend to capture surface-level features and local syntactic information, while semantic and contextual information becomes increasingly salient from the middle to higher layers~\citep{jawahar2019bert}. It has also been observed that information corresponding to different stages of language understanding emerges progressively along the depth of the network~\citep{tenney2019bert}. Building on these observations, prior work has proposed methods that identify which internal computations mediate factual associations and update knowledge by editing those localized computations~\citep{meng2022rome,meng2023memit}. Moreover, as an inference-time attempt to directly intervene in internal representations that arise in layers and attention mechanisms to elicit desirable properties, Inference-Time Intervention (ITI) has been proposed, which improves truthfulness by manipulating the activations of specific attention heads~\citep{li2023inferencetime}.

\paragraph{Contrastive Decoding.}
% Contrastive Decoding（CD）\citep{li2022contrastive}は、expert と amateur の対数確率差を対比目的としつつ、expert の確信度に基づく妥当性制約 $V_{\text{head}}$ で候補を絞ってデコードする手法であり、DoLa \citep{chuang2024dola} はこの対比を単一モデル内の層間に拡張して最終層（mature）と浅い層（premature）の分布を対比し、各トークンで最終層との Jensen--Shannon Divergence（JSD）が最大となる premature 層を動的に選択する。関連研究として、候補トークンごとの層方向の予測確率変化から token-wise cross-layer entropy を算出して最終予測分布を再重み付けするEND \citep{wu2025end} や、RAG において文脈が不確実性（エントロピー）を低減する度合い等に基づきコントラスト強度を適応的に調整するACD \citep{kim2024acd}がある。一方、attention 分布の構造そのものを動的シグナルとして直接用いる設計は限定的である。

Contrastive Decoding (CD)~\citep{li2022contrastive} is a decoding method that uses the log-probability gap between an expert and an amateur as a contrastive objective while restricting candidates with an expert-confidence-based plausibility constraint $V_{\text{head}}$; DoLa~\citep{chuang2024dola} extends this contrastive idea to inter-layer contrast within a single model by contrasting the distributions of the final (mature) layer and a shallow (premature) layer, dynamically selecting at each token the premature layer that maximizes the Jensen--Shannon divergence (JSD) from the final layer. Related work includes END~\citep{wu2025end}, which computes token-wise cross-layer entropy from layer-wise changes in predicted probabilities for candidate tokens and reweights the final predictive distribution accordingly, and ACD~\citep{kim2024acd}, which adaptively adjusts contrast strength in RAG based on, e.g., the extent to which the retrieved context reduces uncertainty (entropy). In contrast, designs that directly exploit the structural patterns of attention distributions as dynamic signals remain limited.

\paragraph{Attention-based Analysis.}
% 注意機構は、入力系列内のどのトークンがどの位置の予測に重要であるかを直接表現するため、LLM の内部動作を理解する上で重要な対象となってきた。これまでの研究では、特定の attention head が構文的依存関係や共参照を符号化すること~\citep{clark2019bert,voita2019analyzing}や、事実知識の参照に関わる head の存在が報告されている~\citep{geva2023dissecting}。また、attention パターンを用いたハルシネーション検出も提案されている ~\citep{chuang2024lookback}。一方で、attention を層選択のシグナルとして動的デコーディングに組み込む試みは限定的である。本研究は DoLa の枠組みを出発点として、注意分布とそのエントロピーを用いた層選択を体系的に評価する点に特徴がある。

Attention mechanisms assign weights to reference tokens when updating representations at each position, and have been used as a cue for observing layer- and head- wise behaviors. Prior work has reported that certain attention heads capture syntactic dependencies and coreference relations \citep{clark2019bert,voita2019analyzing}. It has also been suggested that some heads are involved in factual knowledge recall and attribute extraction \citep{geva2023dissecting}, and that attention patterns can be used to detect hallucinations during generation \citep{chuang2024lookback}. These findings suggest that attention distributions may serve as a signal for layer selection. However, to the best of our knowledge, attempts to directly use attention distributions (or their summaries) for token-level layer selection in dynamic decoding have not been investigated. Starting from the DoLa framework, this study is characterized by a systematic evaluation of layer selection strategies based on attention distributions and their entropy.

\section{Methods}
\label{sec:methods}

In this section, we first review the background of DoLa and then introduce our attention-based layer selection strategies. Finally, we describe the head-level analysis setup.

\subsection{Background: Decoding by Contrasting Layers}
\label{sec:methods-dola}

Given a prefix token sequence $x_{<t}$, a Transformer-based LLM induces a next-token conditional distribution $q_l(x_t \mid x_{<t})$ from the output of each layer $l$.

DoLa \citep{chuang2024dola} is an inference-time decoding method that improves factuality without additional training by contrasting a shallow premature layer with a deep mature layer (the final layer $L$) within the same model. It defines $\hat{P}(x_t \mid x_{<t})$ using a score $F(q_L(x_t), q_{l^*}(x_t))$ based on the ratio between the final-layer distribution and the token-wise selected premature-layer distribution.

\begin{equation} \begin{aligned} \hat{P}(x_t \mid x_{<t}) &= \mathrm{softmax}\!\big(F(q_L(x_t), q_{l^*}(x_t))\big). \end{aligned} \end{equation} \begin{small} \begin{equation} \begin{aligned} F(q_L(x_t), q_{l^*}(x_t)) &= \begin{cases} \log \dfrac{q_L(x_t)}{q_{l^*}(x_t)}, & x_t \in V_{\text{head}}(x_{<t}),\\ -\infty, & \text{otherwise}. \end{cases} \end{aligned} \end{equation} \end{small}

To ensure plausibility under the final layer, candidates are restricted to the token set $V_{\mathrm{head}}(x_{<t})$. Here, $V_{\mathrm{head}}(x_{<t})$ is defined as follows, where $w$ ranges over the vocabulary $\mathcal{V}$:

\begin{equation}
V_{\mathrm{head}}(x_{<t}) =
\left\{
x_t : q_L(x_t) \geq \alpha \max_{w \in \mathcal{V}} q_L(w)
\right\}.
\end{equation}

The premature layer is then selected dynamically at each token by maximizing the Jensen--Shannon Divergence (JSD) from the final layer over a candidate set $\mathcal{C}$. Here, $\mathcal{C}$ denotes the set of candidate premature layers used for contrast; in DoLa, layers are partitioned into several buckets (contiguous ranges), one bucket is chosen via validation, and the layers within the selected bucket (including a setting that considers only even-numbered layers for efficiency) constitute $\mathcal{C}$.

\begin{equation}
l^* =
\operatorname*{arg\,max}_{l \in \mathcal{C}}
\operatorname{JSD}\left(
q_L(\cdot \mid x_{<t}),
q_l(\cdot \mid x_{<t})
\right).
\end{equation}

\subsection{Attention-Guided Layer Selection}
\label{sec:methods-attention}

DoLa selects premature layers based on divergences in token (vocabulary) distributions, but this criterion does not directly capture the underlying reference structure (i.e., where the model attends when making predictions). To address this limitation, we use attention distributions as a layer-selection signal that reflects layer-wise differences in which the model attends, and introduce three attention-guided layer selection strategies while preserving the DoLa framework.

Let $A^l \in \mathbb{R}^{H \times T \times T}$ denote the self-attention tensor at layer $l$, where $A^l_{h,t,i}$ is the attention weight assigned by head $h$ at query position $t$ to position $i$ (a past token). For query position $t$, we define the head-averaged attention row over past tokens $1:t$ as $a^l_t \in \mathbb{R}^{t}$:

\begin{equation}
\begin{aligned}
a^l_t(i)
&= \frac{1}{H}\sum_{h=1}^{H} A^l_{h,t,i},
\qquad i \in \{1,\ldots,t\}, \\
a^l_t(i) &\geq 0,
\qquad
\sum_{i=1}^{t} a^l_t(i) = 1.
\end{aligned}
\end{equation}

\subsubsection{Attention-JSD}
\label{sec:methods-attn-jsd}

Since attention distributions can reflect where each layer attends when making predictions, a layer whose attention pattern differs substantially from the final layer may correspond to a different stage of computation. Under this hypothesis, Attention-JSD selects the layer whose attention distribution is most divergent from the final layer:

\begin{equation}
l^*_{\mathrm{attn\text{-}jsd}} =
\operatorname*{arg\,max}_{l \in \mathcal{C}}
\operatorname{JSD}\left(a^L_t, a^l_t\right).
\end{equation}

\subsubsection{Attention-Entropy-Max}
\label{sec:methods-attn-ent-max}

The entropy of an attention distribution can be interpreted as a summary of how broadly attention is spread across past tokens. Under the hypothesis that broadly distributed attention can serve as a cue for layer selection, Attention-Entropy-Max selects the layer with the maximum attention entropy:

\begin{equation}
\begin{aligned}
H(a^l_t)
&= -\sum_{i=1}^{t} a^l_t(i)\log a^l_t(i), \\
l^*_{\mathrm{attn\text{-}ent\text{-}max}}
&= \operatorname*{arg\,max}_{l \in \mathcal{C}} H(a^l_t).
\end{aligned}
\end{equation}

\subsubsection{Attention-Entropy-Min}
\label{sec:methods-attn-ent-min}

Conversely, low-entropy attention indicates that attention is concentrated on a small number of tokens, which may reflect strong reliance on specific evidence tokens. Under the hypothesis that such concentrated attention can serve as a cue for layer selection, Attention-Entropy-Min selects the layer with the minimum attention entropy:

\begin{equation}
l^*_{\mathrm{attn\text{-}ent\text{-}min}} =
\operatorname*{arg\,min}_{l \in \mathcal{C}} H(a^l_t).
\end{equation}

\subsection{Head-wise Attention Analysis}
\label{sec:methods-head}

To better understand the behavior of our attention-based methods, we conduct a head-wise analysis. 

In the definitions above, the attention distribution $a^l_t$ is obtained by averaging over all $H$ heads. In this analysis, we vary the range of heads used for averaging, and apply the resulting attention distributions to the computations of Attention-JSD and Attention-Entropy. We compare the following settings:

\begin{itemize}
    \item \textbf{All heads}: Use all heads (default).
    \item \textbf{First-half}: Use only the first half of heads ($1$ to $H/2$).
    \item \textbf{Second-half}: Use only the second half of heads ($H/2 + 1$ to $H$), assuming $H$ is even.
\end{itemize}

\section{Experimental Setup}
\label{sec:setup}

\subsection{Models}
\label{sec:setup-models}

In this study, we primarily report results on six representative models: LLaMA-7B\footnote{\url{https://huggingface.co/huggyllama/llama-7b}}, LLaMA-13B\footnote{\url{https://huggingface.co/huggyllama/llama-13b}}, LLaMA-33B\footnote{\url{https://huggingface.co/huggyllama/llama-30b}}, LLaMA-65B\footnote{\url{https://huggingface.co/huggyllama/llama-65b}} for scaling analysis within the LLaMA family, plus Gemma-7B\footnote{\url{https://huggingface.co/google/gemma-7b}} and Mistral-7B\footnote{\url{https://huggingface.co/mistralai/Mistral-7B-v0.1}} as representative non-LLaMA architectures.

Notably, LLaMA-33B and LLaMA-65B were executed with int8 quantization due to hardware memory constraints, while the other models were run in their native precision. Additional model details are provided in Appendix~\ref{sec:appendix-additional-model-details}. 

\begin{table}[t]
\centering
\caption{
Summary of the six representative models used in the main tables. Layers and Heads denote the numbers of Transformer blocks and attention heads, respectively.
}
\label{tab:models-main}
\small
\begin{tabular}{lrcc}
\toprule
\textbf{Model} & \textbf{Params} & \textbf{Layers} & \textbf{Heads} \\
\midrule
LLaMA-7B    & 7B  & 32 & 32 \\
LLaMA-13B   & 13B & 40 & 40 \\
LLaMA-33B$^\dagger$ & 33B & 60 & 52 \\
LLaMA-65B$^\dagger$ & 65B & 80 & 64 \\
Gemma-7B    & 7B  & 28 & 16 \\
Mistral-7B  & 7B  & 32 & 32 \\
\bottomrule
\end{tabular}

\vspace{0.5em}
\begin{minipage}{0.8\linewidth}
\footnotesize{$^\dagger$Executed with int8 quantization.}
\end{minipage}
\end{table}

\subsection{Datasets}
\label{sec:setup-datasets}

We use two representative benchmarks to evaluate the factuality of LLMs. For both benchmarks, we follow the same evaluation protocol as DoLa \citep{chuang2024dola}.

\paragraph{TruthfulQA.}
TruthfulQA is a benchmark designed to measure the tendency of LLMs to generate incorrect answers rooted in misconceptions or myths. We use the multiple-choice setting and report the following three metrics:
\begin{itemize}
  \item \textbf{MC1}: Accuracy in the setting where only a single correct option exists.
  \item \textbf{MC2}: Normalized score in the setting where multiple correct options exist.
  \item \textbf{MC3}: Unnormalized score in the setting where multiple correct options exist.
\end{itemize}

\paragraph{FACTOR.}
FACTOR is a benchmark for evaluating factuality in long-form text. Each instance consists of a prefix and four candidate continuations, only one of which is factually correct. The dataset contains 4,266 four-choice questions spanning three domains: Wiki (2,994), News (1,036), and Expert (236). In this study, we evaluate the Wiki and News domains, using multiple-choice scoring by comparing the log-likelihoods of candidate continuations.

\subsection{Implementation Details}
\label{sec:setup-impl}

In DoLa, the candidate set of premature layers $\mathcal{C}$ is not taken directly from all layers; instead, layers are partitioned into several contiguous ranges (buckets), and the layers within a single bucket selected via validation are used as candidates (i.e., $\mathcal{C}$). In this section, we describe the bucket partition settings (candidate buckets).

\paragraph{Candidate Buckets.}
For the LLaMA family (7B, 13B, 33B, and 65B) and Mistral-7B, we follow the layer-selection ranges (bucket partitions) defined in the original DoLa paper \citep{chuang2024dola}. Specifically, for LLaMA-7B (32 layers) and Mistral-7B (32 layers), we use two buckets: $[0, 16)$ and $[16, 32)$. For LLaMA-13B (40 layers), we use two buckets: $[0, 20)$ and $[20, 40)$. For LLaMA-33B (60 layers), we use three buckets: $[0, 20)$, $[20, 40)$, and $[40, 60)$. For LLaMA-65B (80 layers), we use four buckets: $[0, 20)$, $[20, 40)$, $[40, 60)$, and $[60, 80)$.

In contrast, for Gemma-7B (28 layers), we partition the layers into two buckets: $[0, 14)$ and $[14, 28)$. Notably, for Gemma-7B, we exclude layer 0 from the candidate set for all methods, including the DoLa baseline (see Appendix~\ref{sec:appendix-layer0}), as it primarily represents non-semantic embedding information. Furthermore, within the selected bucket, we only consider even-indexed layers as candidates to reduce computational overhead.

\paragraph{Bucket Selection.}
For TruthfulQA, following DoLa \citep{chuang2024dola}, we perform 2-fold validation and choose the bucket that yields the highest MC3 score. For FACTOR, we treat the Wiki and News subsets as two independent folds and select the optimal bucket for each. Table~\ref{tab:layer-ranges} reports the buckets selected by Attention-JSD for the representative models.

\begin{table}[t]
\centering
\caption{
Selected premature-layer ranges for Attention-JSD across representative models.
}
\label{tab:layer-ranges}
\small
\begin{tabular}{l|cc|c}
\toprule
\textbf{Model} & \textbf{TruthfulQA} & \textbf{FACTOR} & \textbf{Mature} \\
\midrule
LLaMA-7B    & $[16, 32)$ & $[2, 16)$ & 32 \\
LLaMA-13B   & $[20, 40)$ & $[2, 20)$ & 40 \\
LLaMA-33B   & $[20, 40)$ & $[2, 20)$ & 60 \\
LLaMA-65B   & $[60, 80)$ & $[2, 20)$ & 80 \\
Gemma-7B    & $[14, 28)$ & $[14, 28)$ & 28 \\
Mistral-7B  & $[16, 32)$ & $[2, 16)$ & 32 \\
\bottomrule
\end{tabular}
\end{table}

\section{Results}
\label{sec:results}

This section reports results for the six representative models: LLaMA-7B, 13B, 33B, 65B, Gemma-7B, and Mistral-7B. Results for additional models are provided in Appendix~\ref{sec:appendix-full-results}.

\subsection{TruthfulQA}
\label{sec:results-truthfulqa}

Table~\ref{tab:truthfulqa-results} reports the results of the six representative models on TruthfulQA. Here, the Baseline refers to evaluating each model with standard autoregressive decoding. The three proposed attention-distribution-based strategies consistently outperformed the Baseline across all evaluated models. Notably, the attention-based methods (Attention-JSD and Attention-Entropy-Max) contributed substantially to improving MC2 and MC3, which consider multiple acceptable answers.

Specifically, for LLaMA-7B, Attn-Ent-Min achieved the highest performance in MC2 (62.8\%) and MC3 (36.5\%), significantly surpassing DoLa. In LLaMA-13B, Attn-JSD yielded the best results across all metrics. For LLaMA-33B and LLaMA-65B, while DoLa maintained the highest MC1, our attention-based methods outperformed DoLa in both MC2 and MC3. This suggests that internal attention distributions provide a more sensitive signal than output vocabulary distributions for extracting factual knowledge, regardless of model scale.

A similar trend was observed in Gemma-7B and Mistral-7B, where the proposed methods consistently maintained or improved upon DoLa's scores in the MC2 and MC3 metrics. Overall, these results confirm that utilizing structural information from attention distributions, such as attention concentration (Min-Entropy) and divergence (JSD), serves as a robust mechanism for dynamically selecting contrastive layers to enhance factual accuracy.

\begin{table*}[t]
\centering
\caption{
Results on TruthfulQA for the six representative models (MC1 / MC2 / MC3, \%). \textbf{Bold} indicates the best result for each model.
}
\label{tab:truthfulqa-results}
\small
\begin{tabular}{l|ccccc}
\toprule
\textbf{Model} & \textbf{Baseline} & \textbf{DoLa} & \textbf{Attn-JSD} & \textbf{Attn-Ent-Max} & \textbf{Attn-Ent-Min} \\
\midrule
LLaMA-7B 
  & 25.3 / 40.4 / 20.6 
  & \textbf{34.6} / 56.5 / 30.5 
  & 34.1 / 56.2 / 29.3 
  & 34.0 / 55.4 / 28.8 
  & 33.4 / \textbf{62.8} / \textbf{36.5} \\
LLaMA-13B 
  & 28.3 / 42.7 / 22.4 
  & 30.5 / 59.0 / 32.9  
  & \textbf{32.2} / \textbf{61.5} / \textbf{36.0}  
  & 32.1 / 60.4 / 35.4  
  & 30.2 / 60.7 / 34.2  \\
LLaMA-33B 
  & 30.4 / 47.1 / 24.6 
  & \textbf{32.0} / 57.7 / 33.0 
  & 30.8 / 57.5 / 33.1 
  & 29.0 / \textbf{57.8} / \textbf{33.3} 
  & 30.4 / 57.6 / 32.8  \\
LLaMA-65B 
  & 30.8 / 46.5 / 24.8 
  & \textbf{33.3} / 58.7 / 32.7 
  & 32.3 / 59.0 / 33.8 
  & 31.0 / \textbf{59.7} / \textbf{34.6} 
  & 32.4 / 59.2 / 33.9  \\
Gemma-7B 
  & 31.5 / 47.5 / 24.6 
  & 36.8 / 57.7 / 31.1 
  & 36.8 / 57.7 / \textbf{31.2} 
  & 36.5 / 57.6 / 30.9 
  & \textbf{37.2} / \textbf{58.2} / 31.1  \\
Mistral-7B 
  & 30.8 / 48.0 / 25.6 
  & \textbf{35.1} / 56.3 / 29.9 
  & 35.0 / \textbf{56.4} / \textbf{30.1} 
  & 35.0 / 56.3 / 29.9 
  & 35.0 / 56.3 / 29.9  \\
\bottomrule
\end{tabular}
\end{table*}

\subsection{FACTOR}
\label{sec:results-factor}

Table~\ref{tab:factor-results} presents the evaluation results on the FACTOR benchmark. Overall, our proposed attention-guided strategies consistently improved factual accuracy over the Baseline. Furthermore, our methods demonstrated performance comparable to DoLa, with several configurations showing additional improvements in identifying factually correct continuations.

Specifically, on the LLaMA-7B Wiki subset, Attn-Ent-Max achieved an accuracy of 62.6\%, outperforming DoLa (62.1\%). For the Gemma-7B Wiki subset, Attn-JSD reached 64.1\%, the highest among all compared methods, surpassing DoLa's 63.3\%. In the Mistral-7B News subset, Attn-Ent-Max yielded the best performance with 76.3\% accuracy.

While attention-guided methods did not surpass DoLa in certain cases, such as the Gemma-7B News subset, our results across most configurations indicate that attention-based signals---specifically attention entropy and JSD---serve as a powerful and competitive alternative to logit-based signals for identifying layers rich in factual knowledge.

\begin{table*}[t]
\centering
\caption{
Results on FACTOR for the six representative models (Accuracy, \%). \textbf{Bold} indicates the best result per model and split. \underline{Underline} indicates a result worse than the Baseline.
}
\label{tab:factor-results}
\small
\begin{tabular}{ll|ccccc}
\toprule
\textbf{Model} & \textbf{Split} & \textbf{Baseline} & \textbf{DoLa} & \textbf{Attn-JSD} & \textbf{Attn-Ent-Max} & \textbf{Attn-Ent-Min} \\
\midrule
\multirow{2}{*}{LLaMA-7B}
  & Wiki & 58.3 & 62.1 & 62.4 & \textbf{62.6} & 62.0 \\
  & News & 58.2 & 61.7 & 61.5 & \textbf{62.0} & \textbf{62.0} \\
\midrule
\multirow{2}{*}{LLaMA-13B}
  & Wiki & 62.5 & 66.2 & 66.3 & \textbf{66.5} & 66.2 \\
  & News & 60.7 & 62.5 & 62.7 & \textbf{63.4} & 63.0 \\
\midrule
\multirow{2}{*}{LLaMA-33B}
  & Wiki & 68.3 & \textbf{69.0} & 68.6 & \textbf{69.0} & 68.8 \\
  & News & 62.5 & 63.5 & 65.3 & 65.2 & \textbf{65.5} \\
\midrule
\multirow{2}{*}{LLaMA-65B}
  & Wiki & 72.1 & \underline{70.4} & \underline{70.9} & \underline{70.8} & \underline{71.1} \\
  & News & 62.8 & 63.5 & 65.2 & \textbf{65.3} & 65.2 \\
\midrule
\multirow{2}{*}{Gemma-7B}
  & Wiki & 60.5 & 63.3 & \textbf{64.1} & 63.3 & 63.2 \\
  & News & 74.0 & \textbf{74.8} & \underline{73.3} & \underline{73.6} & \underline{72.7} \\
\midrule
\multirow{2}{*}{Mistral-7B}
  & Wiki & 60.6 & \textbf{64.6} & 64.5 & 64.4 & 64.4 \\
  & News & 75.9 & 75.9 & \underline{75.3} & \textbf{76.3} & 76.0 \\
\bottomrule
\end{tabular}
\end{table*}

\section{Analysis}
\label{sec:analysis}

\subsection{Layer Selection Patterns}
\label{sec:analysis-layer-selection}

\begin{figure}[t]
  \centering
  \includegraphics[width=0.98\linewidth]{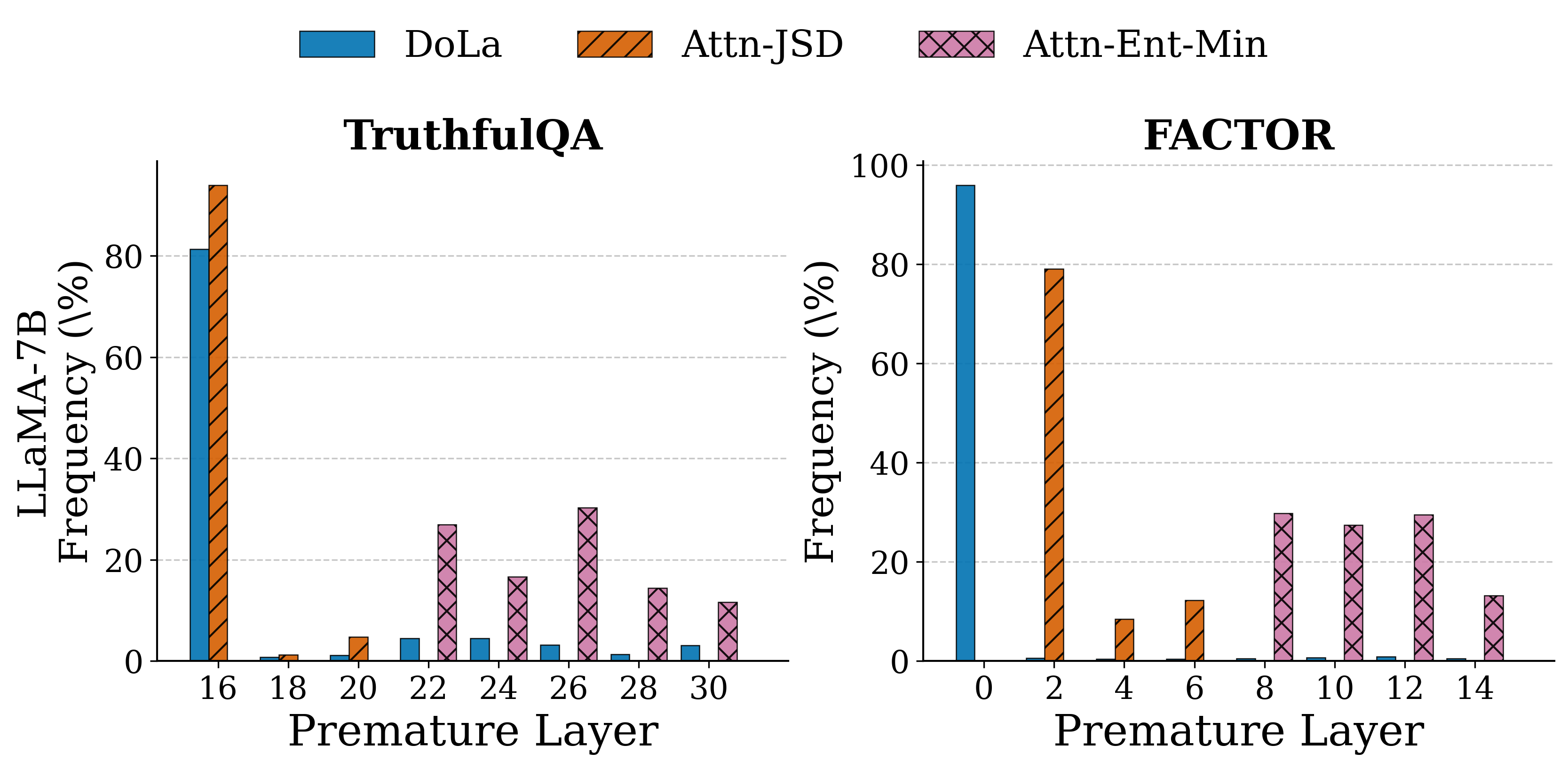}
  \caption{
    Layer selection distributions for LLaMA-7B on TruthfulQA (left) and FACTOR Wiki (right).
  }
  \label{fig:layer-selection-llama7b}
\end{figure}

Figure~\ref{fig:layer-selection-llama7b} illustrates the layer selection distributions for each method on LLaMA-7B. Here, we focus on methods that exhibit noticeable differences in performance (DoLa, Attn-JSD, and Attn-Ent-Min). Both DoLa and Attn-JSD exhibit a strong tendency to concentrate their selections on the shallowest available layers within the candidate bucket. On TruthfulQA, both methods select layer 16 in over 80\% of cases.

In contrast, Attn-Ent-Min demonstrates a markedly different behavior, with selections distributed across a wider range of relatively deeper layers. On TruthfulQA, selections are spread across layers 22--30, and on FACTOR Wiki, they are distributed throughout the middle layers (8--14). This pattern indicates that the layer with the minimum attention entropy varies dynamically for each token, suggesting that Attn-Ent-Min adaptively selects premature layers based on the specific input context.

This qualitative difference in layer selection patterns suggests a potential contribution to the improvements observed in the MC2 and MC3 metrics, which evaluate the model's performance across multiple correct answer options.

\subsection{Head-wise Analysis}
\label{sec:analysis-headwise}

To investigate which attention heads contribute to layer selection in attention-based methods, we conducted a detailed head-wise analysis on TruthfulQA.

\paragraph{Coarse-grained Analysis.}
Table~\ref{tab:headwise} presents the performance of Attn-Ent-Min across different head configurations. We observe that the optimal head group for layer selection varies by model. For LLaMA-7B, the First-half configuration achieved the best performance across all metrics, surpassing both the All-heads average and the Second-half group.

Conversely, for Gemma-7B, utilizing all heads resulted in the highest performance. In the case of Mistral-7B, the scores remained identical across all head configurations.

These results suggest that the distribution of layer-selection signals within attention entropy depends on the model architecture. While informative signals may be concentrated in specific head subsets for some models, they appear to be more dispersed or uniformly distributed in others.

\paragraph{Single-Head Evaluation.}
To further investigate the characteristics of the signals within individual attention heads, we analyzed their layer-selection tendencies when using each head alone. Specifically, we compared each head's selected layers against the all-heads average selection (All heads avg.) and focused on heads that most strongly deviate from this aggregate behavior. We selected Heads 6, 24, and 26 as the top three heads with the lowest agreement with the all-heads average selection. Table~\ref{tab:single-head} presents their evaluation results on LLaMA-7B.

The results show that decoding guided solely by Head 6 achieved an MC3 score of 38.4\%, significantly exceeding both DoLa (30.5\%) and the all-heads average (36.8\%). While the all-heads average selection provides a stable and robust baseline, combining these top three heads also maintained high performance, consistently surpassing both DoLa and the All heads avg. configuration.

These findings demonstrate that beyond simple averaging, factuality can be further improved by isolating and utilizing specific internal signals based on the quality and characteristics of the information they carry.

\begin{table}[t]
\centering
\caption{
Head-wise analysis for Attn-Ent-Min on TruthfulQA. Best result per model is in \textbf{bold}.
}
\label{tab:headwise}
\small
\begin{tabular}{llccc}
\toprule
\textbf{Model} & \textbf{Heads} & \textbf{MC1} & \textbf{MC2} & \textbf{MC3} \\
\midrule
\multirow{3}{*}{LLaMA-7B}
  & All         & \textbf{33.4} & 62.8 & 36.5 \\
  & First-half  & \textbf{33.4} & \textbf{62.9} & \textbf{37.3} \\
  & Second-half & 32.1 & 62.2 & 36.3 \\
\midrule
\multirow{3}{*}{Gemma-7B}
  & All         & \textbf{37.2} & \textbf{58.2} & \textbf{31.1} \\
  & First-half  & 36.2 & 57.5 & 30.5 \\
  & Second-half & 36.8 & 58.0 & 31.0 \\
\midrule
\multirow{3}{*}{Mistral-7B}
  & All         & \textbf{35.0} & \textbf{56.3} & \textbf{29.9} \\
  & First-half  & \textbf{35.0} & \textbf{56.3} & \textbf{29.9} \\
  & Second-half & \textbf{35.0} & \textbf{56.3} & \textbf{29.9} \\
\bottomrule
\end{tabular}
\end{table}

\begin{table}[t]
\centering
\caption{
Single-head evaluation for Attn-Ent-Min on TruthfulQA (LLaMA-7B). Best result per metric is \textbf{bold}.
}
\label{tab:single-head}
\small
\begin{tabular}{lccc}
\toprule
\textbf{Configuration} & \textbf{MC1} & \textbf{MC2} & \textbf{MC3} \\
\midrule
DoLa                  & 34.6 & 56.5 & 30.5 \\
All heads avg. (32)   & 32.4 & 62.7 & 36.8 \\
\midrule
Head 6                & \textbf{35.3} & \textbf{64.9} & \textbf{38.4} \\
Heads 6, 24, 26       & 34.8 & 63.9 & 38.1 \\
\bottomrule
\end{tabular}
\end{table}

\subsection{Head Stability Analysis}
\label{sec:analysis-head-stability}

This section provides detailed results of the head stability analysis. We investigated whether the factuality-improving effects of specific attention heads are localized phenomena dependent on certain question sets or if they represent statistically stable characteristics across different contexts.

We generated five random subsets of 100 questions each from TruthfulQA and evaluated all 32 heads individually using the Attn-Ent-Min strategy on LLaMA-7B.

Table~\ref{tab:head-stability} lists the top-3 performing heads for each random subset based on the MC3 score.

As a result, Head 6 achieved the highest MC3 score in three out of the five subsets. While the performance ranking fluctuates depending on the subset, these results suggest a strong tendency for specific internal signals to be frequently associated with higher factual accuracy. This indicates that it may be possible to pre-identify or estimate superior configurations for layer selection---such as specific attention heads---to further optimize decoding performance.

\begin{table}[t]
\centering
\caption{
Top-3 performing heads for Attn-Ent-Min on each random subset (LLaMA-7B, MC3).
}
\label{tab:head-stability}
\small
\begin{tabular}{lccc}
\toprule
\textbf{Subset} & \textbf{Rank 1} & \textbf{Rank 2} & \textbf{Rank 3} \\
\midrule
Subset 1 & \textbf{Head 6}  & Head 1  & Head 22 \\
Subset 2 & Head 1  & Head 15 & Head 10 \\
Subset 3 & Head 26 & \textbf{Head 6}  & Head 5  \\
Subset 4 & \textbf{Head 6}  & Head 22 & Head 26 \\
Subset 5 & \textbf{Head 6}  & Head 26 & Head 22 \\
\bottomrule
\end{tabular}
\end{table}

\subsection{Score Distribution Analysis}
\label{sec:analysis-score-distribution}

On LLaMA-33B, DoLa achieved the highest MC1 score (31.9\%), while Attention-Entropy-Max showed a lower MC1 (29.0\%) but slightly outperformed DoLa in MC2 and MC3 (MC2: 57.8\% vs 57.7\%, MC3: 33.3\% vs 33.0\%). To understand this difference, we analyzed the score distributions of individual samples when using DoLa and attention-based methods.

We first classified samples into four categories based on MC1 correctness (Table~\ref{tab:mc1-correctness-patterns}). Among the three attention-based methods, Attn-Ent-Max shows the least overlap with DoLa: only 150 samples are correct under both methods, while 111 are correct only with DoLa and 87 only with Attn-Ent-Max. This indicates that the two methods tend to correctly answer different samples. In contrast, Attn-JSD and Attn-Ent-Min show greater overlap with DoLa (191--194 samples correct under both), and their MC1 scores are closer to DoLa's.

We then analyzed the overall score distribution characteristics of each method (Table~\ref{tab:score-distribution-stats}). For top margin (the gap between the highest correct and incorrect scores), DoLa achieves the largest value (14.8), while attention-based methods show smaller values (8.5--10.3). Notably, Attn-Ent-Max has the smallest top margin (8.5), indicating the weakest discriminability for identifying the top correct answer. Conversely, for true std (standard deviation of correct option scores), DoLa shows the largest value (31.3), while Attn-Ent-Max shows the smallest (17.5). This suggests that DoLa generates ``sharp'' distributions concentrating confidence on specific correct answers, whereas Attn-Ent-Max generates ``flat'' distributions spreading scores more evenly across all correct options.

These results reveal that DoLa's sharp layer selection is effective for identifying a single best answer (MC1), while attention-based methods with flatter selection patterns are advantageous for comprehensively evaluating multiple correct answers (MC2/MC3), demonstrating complementary characteristics between the two approaches.

\begin{table}[t]
\centering
\caption{
Sample-level correctness patterns (MC1) on TruthfulQA for LLaMA-33B.
}
\label{tab:mc1-correctness-patterns}
\small
\setlength{\tabcolsep}{3pt}
\begin{tabular}{l|cccc}
\toprule
\textbf{Comparison} & \textbf{Both} & \textbf{DoLa} & \textbf{Attn} & \textbf{Both} \\
                    & \textbf{corr.} & \textbf{only} & \textbf{only} & \textbf{wrong} \\
\midrule
DoLa vs Attn-JSD     & 194 & 67  & 58 & 498 \\
DoLa vs Attn-Ent-Max & 150 & 111 & 87 & 469 \\
DoLa vs Attn-Ent-Min & 191 & 70  & 57 & 499 \\
\bottomrule
\end{tabular}
\end{table}

\begin{table}[t]
\centering
\caption{
Overall score distribution statistics (LLaMA-33B). Top margin = max(correct) $-$ max(incorrect). True std = standard deviation among correct option scores.
}
\label{tab:score-distribution-stats}
\small
\begin{tabular}{lcc}
\toprule
\textbf{Method} & \textbf{Top Margin} & \textbf{True Std} \\
\midrule
DoLa             & 14.8 & 31.3 \\
Attn-JSD         & 10.3 & 23.6 \\
Attn-Ent-Max     & 8.5  & 17.5 \\
Attn-Ent-Min     & 10.2 & 23.1 \\
\bottomrule
\end{tabular}
\end{table}

\section{Conclusion}
\label{sec:conclusion}

In this work, we proposed three strategies that leverage self-attention mechanisms rather than conventional vocabulary distributions as dynamic signals for layer selection in contrastive decoding. Our evaluation on TruthfulQA and FACTOR demonstrates that the proposed methods achieve factuality improvements equal to or greater than the existing DoLa method across multiple models.

In particular, our score distribution analysis suggests that DoLa tends to produce sharper distributions that concentrate confidence on a specific correct answer, which is effective for MC1, whereas attention-based methods produce flatter distributions that spread scores across all correct options, which is effective for MC2/MC3.

Furthermore, our head-wise analysis confirms that some attention heads exhibit stronger factuality signals than the all-head average. This suggests that further improvements in factuality and efficiency may be achieved by context-dependent head selection or by optimizing the set of heads used. In addition, our method requires no additional training and keeps the inference-time computational overhead within a practical range. 

As future work, we will evaluate our approach in free-form generation settings and explore automated, context-dependent selection of layers and heads.

\section*{Limitations}

This work has several limitations that define important directions for future research.

First, our evaluation is primarily limited to multiple-choice benchmarks. A crucial next step is to verify how attention-guided layer selection functions in free-form generation tasks, where sequential token dependencies may differ from likelihood based ranking.

Second, the selection of the candidate premature-layer bucket still requires prior validation. While attention-based signals provide sensitive cues, developing algorithms that can fully automate bucket selection or generalize across different model depths remains a necessary improvement.

Third, the underlying mechanism of why specific attention heads excel at factual knowledge recall remains to be fully elucidated. While the statistical stability observed in this study suggests that certain internal representations mediate truthfulness, identifying the theoretical basis for this localization is an essential task for future investigation.

\bibliography{custom}

\newpage
\appendix

\section{Additional Model Details}
\label{sec:appendix-additional-model-details}

In addition to the six representative models reported in the main text, we also ran experiments on four additional models. In this appendix, we additionally report results for these models (LLaMA-3.1-8B\footnote{\url{https://huggingface.co/meta-llama/Llama-3.1-8B}}, LLaMA-3.2-3B\footnote{\url{https://huggingface.co/meta-llama/Llama-3.2-3B}}, Gemma-2-2B\footnote{\url{https://huggingface.co/google/gemma-2-2b}}, and Phi-2\footnote{\url{https://huggingface.co/microsoft/phi-2}}). Their specifications are summarized in Table~\ref{tab:appendix-additional-models}.

\begin{table}[h]
\centering
\caption{
Additional models used in this study.
}
\label{tab:appendix-additional-models}
\small
\begin{tabular}{lrcc}
\toprule
\textbf{Model} & \textbf{Params} & \textbf{Layers} & \textbf{Heads} \\
\midrule
LLaMA-3.1-8B & 8B   & 32 & 32 \\
LLaMA-3.2-3B & 3B   & 28 & 24 \\
Gemma-2-2B   & 2.6B & 26 & 8  \\
Phi-2        & 2.7B & 32 & 32 \\
\bottomrule
\end{tabular}
\end{table}

\subsection{Handling of Layer 0 in Models with Tied Embeddings}
\label{sec:appendix-layer0}

The DoLa paper notes that when the input embeddings and the LM head share weights, the resulting operation becomes close to an identity mapping; therefore, in such cases, layer 0 is excluded from the candidate set. In our implementation, we consult the Hugging Face configuration (e.g., \texttt{tie\_word\_embeddings}) and exclude layer 0 when \texttt{tie\_word\_embeddings} is true (i.e., when weight tying is enabled), applying this rule to models such as Gemma variants and Llama 3.2-3B.

\section{Selected Premature-Layer Ranges by Method}
\label{sec:appendix-layer-ranges}

The range of candidate premature layers for dynamic selection is determined via validation for each combination of model and dataset. Tables~\ref{tab:appendix-ranges-dola} through \ref{tab:appendix-ranges-ent-min} list the selected buckets for all evaluated models across the four contrastive decoding methods.

\begin{table}[h]
\centering
\caption{
Selected premature-layer ranges for DoLa.
}
\label{tab:appendix-ranges-dola}
\small
\begin{tabular}{l|cc|c}
\toprule
\textbf{Model} & \textbf{TruthfulQA} & \textbf{FACTOR} & \textbf{Mature} \\
\midrule
LLaMA-7B     & $[16, 32)$ & $[0, 16)$  & 32 \\
LLaMA-13B    & $[20, 40)$ & $[0, 20)$  & 40 \\
LLaMA-33B    & $[40, 60)$ & $[0, 20)$  & 60 \\
LLaMA-65B    & $[60, 80)$ & $[0, 20)$  & 80 \\
Gemma-7B     & $[14, 28)$ & $[14, 28)$ & 28 \\
Mistral-7B   & $[0, 16)$  & $[0, 16)$  & 32 \\
\midrule
LLaMA-3.1-8B & $[0, 16)$  & $[16, 32)$ & 32 \\
LLaMA-3.2-3B & $[2, 14)$  & $[2, 14)$  & 28 \\
Gemma-2-2B   & $[2, 14)$  & $[2, 14)$  & 26 \\
Phi-2        & $[16, 32)$ & $[0, 16)$  & 32 \\
\bottomrule
\end{tabular}
\end{table}

\begin{table}[h]
\centering
\caption{
Selected premature-layer ranges for Attention-JSD.
}
\label{tab:appendix-ranges-jsd}
\small
\begin{tabular}{l|cc|c}
\toprule
\textbf{Model} & \textbf{TruthfulQA} & \textbf{FACTOR} & \textbf{Mature} \\
\midrule
LLaMA-7B     & $[16, 32)$ & $[2, 16)$  & 32 \\
LLaMA-13B    & $[20, 40)$ & $[2, 20)$  & 40 \\
LLaMA-33B    & $[20, 40)$ & $[2, 20)$  & 60 \\
LLaMA-65B    & $[60, 80)$ & $[2, 20)$  & 80 \\
Gemma-7B     & $[14, 28)$ & $[14, 28)$ & 28 \\
Mistral-7B   & $[16, 32)$ & $[2, 16)$  & 32 \\
\midrule
LLaMA-3.1-8B & $[16, 32)$ & $[2, 16)$  & 32 \\
LLaMA-3.2-3B & $[2, 14)$  & $[2, 14)$  & 28 \\
Gemma-2-2B   & $[2, 14)$  & $[2, 14)$  & 26 \\
Phi-2        & $[2, 16)$  & $[2, 16)$  & 32 \\
\bottomrule
\end{tabular}
\end{table}

\begin{table}[h]
\centering
\caption{
Selected premature-layer ranges for Attention-Entropy-Max.
}
\label{tab:appendix-ranges-ent-max}
\small
\begin{tabular}{l|cc|c}
\toprule
\textbf{Model} & \textbf{TruthfulQA} & \textbf{FACTOR} & \textbf{Mature} \\
\midrule
LLaMA-7B     & $[16, 32)$ & $[2, 16)$  & 32 \\
LLaMA-13B    & $[20, 40)$ & $[2, 20)$  & 40 \\
LLaMA-33B    & $[40, 60)$ & $[2, 20)$  & 60 \\
LLaMA-65B    & $[60, 80)$ & $[2, 20)$  & 80 \\
Gemma-7B     & $[14, 28)$ & $[14, 28)$ & 28 \\
Mistral-7B   & $[2, 16)$  & $[16, 32)$ & 32 \\
\midrule
LLaMA-3.1-8B & $[2, 16)$  & $[2, 16)$  & 32 \\
LLaMA-3.2-3B & $[2, 14)$  & $[2, 14)$  & 28 \\
Gemma-2-2B   & $[14, 26)$ & $[14, 26)$ & 26 \\
Phi-2        & $[16, 32)$ & $[2, 16)$  & 32 \\
\bottomrule
\end{tabular}
\end{table}

\begin{table}[h]
\centering
\caption{
Selected premature-layer ranges for Attention-Entropy-Min.
}
\label{tab:appendix-ranges-ent-min}
\small
\begin{tabular}{l|cc|c}
\toprule
\textbf{Model} & \textbf{TruthfulQA} & \textbf{FACTOR} & \textbf{Mature} \\
\midrule
LLaMA-7B     & $[16, 32)$ & $[2, 16)$  & 32 \\
LLaMA-13B    & $[20, 40)$ & $[2, 20)$  & 40 \\
LLaMA-33B    & $[20, 40)$ & $[2, 20)$  & 60 \\
LLaMA-65B    & $[60, 80)$ & $[2, 20)$  & 80 \\
Gemma-7B     & $[2, 14)$  & $[2, 14)$  & 28 \\
Mistral-7B   & $[2, 16)$  & $[16, 32)$ & 32 \\
\midrule
LLaMA-3.1-8B & $[2, 16)$  & $[2, 16)$  & 32 \\
LLaMA-3.2-3B & $[14, 28)$ & $[2, 14)$  & 28 \\
Gemma-2-2B   & $[14, 26)$ & $[2, 14)$  & 26 \\
Phi-2        & $[16, 32)$ & $[2, 16)$  & 32 \\
\bottomrule
\end{tabular}
\end{table}

\section{Efficiency Analysis}
\label{sec:appendix-efficiency}

Table~\ref{tab:appendix-efficiency} shows the inference efficiency of each method on LLaMA-7B. We sampled 100 questions from TruthfulQA and measured the average processing time per sample.

All methods complete within a single forward pass, with overheads remaining within 1.15--1.27$\times$ relative to the Baseline. DoLa incurs a 1.24$\times$ cost due to per-candidate-layer logits and JSD computations. Among attention-based methods, Attn-Ent-Max/Min is lightweight (as low as 1.15$\times$) because it only requires entropy computation, whereas Attn-JSD is the most expensive (1.27$\times$) due to JSD computation between the final-layer attention distribution and each candidate layer. Overall, these costs remain practical given the factuality improvements.

\begin{table}[!htbp]
\centering
\caption{
Inference efficiency on TruthfulQA (LLaMA-7B, 100 samples).
}
\label{tab:appendix-efficiency}
\small
\begin{tabular}{lcc}
\toprule
\textbf{Method} & \textbf{Time (s/sample)} & \textbf{Relative} \\
\midrule
Baseline       & $0.84 \pm 0.32$ & $1.00\times$ \\
DoLa           & $1.03 \pm 0.44$ & $1.24\times$ \\
Attn-JSD       & $1.06 \pm 0.41$ & $1.27\times$ \\
Attn-Ent-Max   & $0.96 \pm 0.38$ & $1.15\times$ \\
Attn-Ent-Min   & $1.00 \pm 0.34$ & $1.20\times$ \\
\bottomrule
\end{tabular}
\end{table}

\section{Full Experimental Results}
\label{sec:appendix-full-results}

This section reports full experimental results for all models evaluated in this study.

\subsection{TruthfulQA: Full Results}
\label{sec:appendix-truthfulqa-full}

Table~\ref{tab:appendix-truthfulqa} shows the results for all models on TruthfulQA. In newer models such as LLaMA-3.1-8B and LLaMA-3.2-3B, our proposed strategies (specifically Attn-JSD and Attn-Ent-Min) demonstrate stable improvements over the Baseline. Notably, on Phi-2, Attn-Ent-Min achieved an MC3 score of 34.8\%, substantially outperforming DoLa (30.3\%).

\begin{table*}[h]
\centering
\caption{
Results on TruthfulQA for all models (MC1 / MC2 / MC3, \%). Bold indicates best per model. Underline indicates worse than Baseline.
}
\label{tab:appendix-truthfulqa}
\small
\begin{tabular}{l|ccccc}
\toprule
\textbf{Model} & \textbf{Baseline} & \textbf{DoLa} & \textbf{Attn-JSD} & \textbf{Attn-Ent-Max} & \textbf{Attn-Ent-Min} \\
\midrule
LLaMA-7B 
  & 25.3 / 40.4 / 20.6
  & \textbf{34.6} / 56.5 / 30.5
  & 34.1 / 56.2 / 29.3
  & 34.0 / 55.4 / 28.8
  & 33.4 / \textbf{62.8} / \textbf{36.5} \\
LLaMA-13B
  & 28.3 / 42.7 / 22.4
  & 30.5 / 59.0 / 32.9
  & \textbf{32.2} / \textbf{61.5} / \textbf{36.0}
  & 32.1 / 60.4 / 35.4
  & 30.2 / 60.7 / 34.2 \\
LLaMA-33B
  & 30.4 / 47.1 / 24.6
  & \textbf{32.0} / 57.7 / 33.0
  & 30.8 / 57.5 / 33.1
  & \underline{29.0} / \textbf{57.8} / \textbf{33.3}
  & 30.4 / 57.6 / 32.8 \\
LLaMA-65B
  & 30.8 / 46.5 / 24.8
  & \textbf{33.3} / 58.7 / 32.7
  & 32.3 / 59.0 / 33.8
  & 31.0 / \textbf{59.7} / \textbf{34.6}
  & 32.4 / 59.2 / 33.9 \\
Gemma-7B
  & 31.5 / 47.5 / 24.6
  & 36.8 / 57.7 / 31.1
  & 36.8 / 57.7 / \textbf{31.2}
  & 36.5 / 57.6 / 30.9
  & \textbf{37.2} / \textbf{58.2} / 31.1 \\
Mistral-7B
  & 30.8 / 48.0 / 25.6
  & \textbf{35.1} / 56.3 / 29.9
  & 35.0 / \textbf{56.4} / \textbf{30.1}
  & 35.0 / 56.3 / 29.9
  & 35.0 / 56.3 / 29.9 \\
\midrule
LLaMA-3.1-8B
  & 31.6 / 49.1 / 26.3
  & \textbf{35.6} / 57.4 / \textbf{30.6}
  & \textbf{35.6} / \textbf{57.5} / \textbf{30.6}
  & 35.5 / 57.4 / \textbf{30.6}
  & 35.5 / 57.4 / 30.5 \\
LLaMA-3.2-3B
  & 27.5 / 44.5 / 23.9
  & \textbf{34.3} / 55.9 / 29.7
  & 33.8 / 55.9 / 29.5
  & 34.2 / 55.9 / 29.6
  & 34.0 / \textbf{56.0} / \textbf{29.8} \\
Gemma-2-2B
  & 27.1 / 42.6 / 22.0
  & \textbf{34.3} / \underline{3.0} / \textbf{31.1}
  & 30.8 / 46.9 / 27.8
  & 30.0 / \textbf{54.3} / 28.0
  & 28.3 / 53.3 / 30.8 \\
Phi-2
  & 28.5 / 42.8 / 21.9
  & 33.4 / 55.5 / 30.3
  & \textbf{35.0} / 55.1 / 28.2
  & 33.4 / 53.3 / 28.7
  & 29.3 / \textbf{59.5} / \textbf{34.8} \\
\bottomrule
\end{tabular}
\end{table*}

\subsection{FACTOR: Full Results}
\label{sec:appendix-factor-full}

Table~\ref{tab:appendix-factor} summarizes the full FACTOR results. While contrastive decoding generally contributes to factuality in both Wiki and News subsets, we observed certain cases, such as Gemma-2-2B, where contrastive methods led to a performance drop relative to the Baseline.

\begin{table*}[h]
\centering
\caption{
Results on FACTOR for all models (Accuracy, \%). \textbf{Bold} indicates best per model. \underline{Underline} indicates worse than Baseline.
}
\label{tab:appendix-factor}
\small
\begin{tabular}{ll|ccccc}
\toprule
\textbf{Model} & \textbf{Split} & \textbf{Baseline} & \textbf{DoLa} & \textbf{Attn-JSD} & \textbf{Attn-Ent-Max} & \textbf{Attn-Ent-Min} \\
\midrule
\multirow{2}{*}{LLaMA-7B}
  & Wiki & 58.3 & 62.1 & 62.4 & \textbf{62.6} & 62.0 \\
  & News & 58.2 & 61.7 & 61.5 & \textbf{62.0} & \textbf{62.0} \\
\midrule
\multirow{2}{*}{LLaMA-13B}
  & Wiki & 62.5 & 66.2 & 66.3 & \textbf{66.5} & 66.2 \\
  & News & 60.7 & 62.5 & 62.7 & \textbf{63.4} & 63.0 \\
\midrule
\multirow{2}{*}{LLaMA-33B}
  & Wiki & 68.3 & \textbf{69.0} & 68.6 & \textbf{69.0} & 68.8 \\
  & News & 62.5 & 63.5 & 65.3 & 65.2 & \textbf{65.5} \\
\midrule
\multirow{2}{*}{LLaMA-65B}
  & Wiki & \textbf{72.1} & \underline{70.4} & \underline{70.9} & \underline{70.8} & \underline{71.1} \\
  & News & 62.8 & 63.5 & 65.2 & \textbf{65.3} & 65.2 \\
\midrule
\multirow{2}{*}{Gemma-7B}
  & Wiki & 60.5 & 63.3 & \textbf{64.1} & 63.3 & 63.2 \\
  & News & 74.0 & \textbf{74.8} & \underline{73.3} & \underline{73.6} & \underline{72.7} \\
\midrule
\multirow{2}{*}{Mistral-7B}
  & Wiki & 60.6 & \textbf{64.6} & 64.5 & 64.4 & 64.4 \\
  & News & 75.9 & 75.9 & \underline{75.3} & \textbf{76.3} & 76.0 \\
\midrule
\multirow{2}{*}{LLaMA-3.1-8B}
  & Wiki & 63.9 & \textbf{67.4} & \textbf{67.4} & 67.3 & 67.1 \\
  & News & 75.6 & 75.9 & \underline{74.6} & \underline{74.8} & \underline{74.6} \\
\midrule
\multirow{2}{*}{LLaMA-3.2-3B}
  & Wiki & 56.5 & \textbf{61.6} & 61.0 & 61.5 & 60.9 \\
  & News & 68.9 & 69.4 & \textbf{69.5} & 69.4 & \textbf{69.5} \\
\midrule
\multirow{2}{*}{Gemma-2-2B}
  & Wiki & \textbf{52.4} & \underline{40.4} & \underline{40.0} & \underline{40.0} & \underline{45.3} \\
  & News & \textbf{67.7} & \underline{46.5} & \underline{44.0} & \underline{45.2} & \underline{52.7} \\
\midrule
\multirow{2}{*}{Phi-2}
  & Wiki & 54.6 & 58.5 & \textbf{58.8} & 58.6 & 55.1 \\
  & News & 57.8 & \textbf{61.7} & 61.4 & 60.3 & 58.6 \\
\bottomrule
\end{tabular}
\end{table*}

\section{Head-wise Analysis: Full Results}
\label{sec:appendix-headwise}

To supplement the analysis in the main text, we provide detailed performance metrics for different head configurations under the Attn-JSD and Attn-Ent-Max strategies. Tables~\ref{tab:headwise-jsd} and \ref{tab:headwise-max} report these results for LLaMA-7B, Gemma-7B, and Mistral-7B.

\begin{table}[h]
\centering
\caption{
Head-wise analysis for Attn-JSD on TruthfulQA. Best result per model is in \textbf{bold}.
}
\label{tab:headwise-jsd}
\small
\begin{tabular}{llccc}
\toprule
\textbf{Model} & \textbf{Heads} & \textbf{MC1} & \textbf{MC2} & \textbf{MC3} \\
\midrule
\multirow{3}{*}{LLaMA-7B}
  & All         & 34.2 & 56.2 & 29.3 \\
  & First-half  & \textbf{34.4} & 55.4 & 28.9 \\
  & Second-half & 34.1 & \textbf{57.2} & \textbf{30.9} \\
\midrule
\multirow{3}{*}{Gemma-7B}
  & All         & \textbf{36.8} & \textbf{57.7} & \textbf{31.2} \\
  & First-half  & 36.0 & 57.0 & 31.0 \\
  & Second-half & 35.0 & 57.0 & 29.8 \\
\midrule
\multirow{3}{*}{Mistral-7B}
  & All         & \textbf{35.0} & 56.4 & \textbf{30.1} \\
  & First-half  & \textbf{35.0} & \textbf{56.6} & 29.9 \\
  & Second-half & 34.8 & 56.2 & 29.7 \\
\bottomrule
\end{tabular}
\end{table}

\begin{table}[h]
\centering
\caption{
Head-wise analysis for Attn-Ent-Max on TruthfulQA. Best result per model is in \textbf{bold}.
}
\label{tab:headwise-max}
\small
\begin{tabular}{llccc}
\toprule
\textbf{Model} & \textbf{Heads} & \textbf{MC1} & \textbf{MC2} & \textbf{MC3} \\
\midrule
\multirow{3}{*}{LLaMA-7B}
  & All         & \textbf{34.0} & \textbf{55.4} & \textbf{28.8} \\
  & First-half  & 33.9 & 54.7 & 28.4 \\
  & Second-half & 33.9 & 54.7 & 28.4 \\
\midrule
\multirow{3}{*}{Gemma-7B}
  & All         & 36.5 & \textbf{57.6} & 30.9 \\
  & First-half  & \textbf{36.6} & \textbf{57.6} & \textbf{31.0} \\
  & Second-half & 35.6 & 57.5 & 30.4 \\
\midrule
\multirow{3}{*}{Mistral-7B}
  & All         & \textbf{35.0} & \textbf{56.3} & \textbf{29.9} \\
  & First-half  & \textbf{35.0} & \textbf{56.3} & \textbf{29.9} \\
  & Second-half & \textbf{35.0} & \textbf{56.3} & \textbf{29.9} \\
\bottomrule
\end{tabular}
\end{table}

\end{document}